\documentclass[letterpaper, 10 pt, conference]{ieeeconf}  % Comment this line out if you need a4paper

\usepackage[draft, markup=bfit, deletedmarkup=sout, authormarkup=brackets]{changes}
\usepackage{booktabs}
\usepackage{amssymb}
\usepackage{amsmath}
\usepackage{cite}
\usepackage{color}
\usepackage{xcolor}
\usepackage{stackengine}

\usepackage{siunitx}
\PassOptionsToPackage{hyphens}{url}\usepackage{hyperref}
\usepackage{subfig}
\usepackage{dblfloatfix}    % To enable figures at the bottom of page
\usepackage{eurosym}
\usepackage{doi}

\usepackage{svg}

\usepackage{tikz}
\usetikzlibrary{trees,positioning,shapes,shadows,arrows}

\usepackage{multirow}

\usepackage[labelsep=period]{caption}

\definecolor{excelblue}{HTML}{4472C4} 
\definecolor{NewBlue}{HTML}{4472C4} 
\definecolor{NewOrange}{HTML}{ED7D31} 
\definecolor{NewGreen}{HTML}{00B050} 

\colorlet{Changes@Color}{blue}
\definechangesauthor[name={Dennis}, color=green]{DKn}
\definechangesauthor[name={Henning}, color=orange]{HZw}
\definechangesauthor[name={Utku}, color=blue]{UCu}
\definechangesauthor[name={Sami}, color=red]{SHa}

\newcommand{\checkmarksmall}{\hspace{-3px}\checkmark}

\DeclareGraphicsExtensions{.pdf,.jpeg,.png, .svg}
\graphicspath{{graphics/}}

\IEEEoverridecommandlockouts                              % This command is only needed if 
\def\mytitle{Towards Flexible Biolaboratory Automation: Container Taxonomy-Based, 3D-Printed Gripper Fingers*}
\title{\LARGE \bf \mytitle}

\def\myauthor{Henning Zwirnmann, Dennis Knobbe, Utku Culha and Sami Haddadin}
\def\mythanks{Henning Zwirnmann, Dennis Knobbe and Sami Haddadin are with the Chair of Robotics and Systems Intelligence, School of Computation, Information and Technology, Technical University of Munich. All authors are with the Munich Institute of Robotics and Machine Intelligence, Technical University of Munich, 80797 Munich, Germany. {\href{mailto:henning.zwirnmann@tum.de}{\tt\small henning.zwirnmann@tum.de}}}

\author{\myauthor% <-this % stops a space
\thanks{
*We gratefully acknowledge the funding of this work by the Alfried Krupp von Bohlen and Halbach Foundation. We gratefully acknowledge the funding of this work by the Deutsche Forschungsgemeinschaft through the Gottfried Wilhelm Leibniz Programme (award to Sami Haddadin; grant no. HA7372/3-1).}% <-this % stops a space
\thanks{\mythanks}%
}

\usepackage{fancyhdr}
\fancypagestyle{firststyle}
{
   \fancyhf{}
   
\fancyhead[LO,LE]{\small{\textbf{2023 IEEE/RSJ International Conference on Intelligent Robots and Systems (IROS) [Accepted Version]} \\\textbf{October 1-5, 2023. Detroit, USA}}}
\fancyfoot[LO,LE]{\copyright\footnotesize{\ 2023 IEEE. Personal use of this material is permitted. Permission from IEEE must be obtained for all other uses, in any current or future media, including reprinting/republishing this material for advertising or promotional purposes, creating new collective works, for resale or redistribution to servers or lists, or reuse of any copyrighted component of this work in other works. DOI: \href{https://doi.org/10.1109/IROS55552.2023.10342218}{10.1109/IROS55552.2023.10342218}}}
\fancyfoot[RO,RE]{\thepage}
}
\fancypagestyle{elsestyle}
{
\fancyhf{}

\fancyfoot[LO,LE]{\copyright\footnotesize{\ 2023 IEEE. Personal use of this material is permitted. Permission from IEEE must be obtained for all other uses, in any current or future media, including reprinting/republishing this material for advertising or promotional purposes, creating new collective works, for resale or redistribution to servers or lists, or reuse of any copyrighted component of this work in other works. DOI: \href{https://doi.org/10.1109/IROS55552.2023.10342218}{10.1109/IROS55552.2023.10342218}}}
\fancyfoot[RO,RE]{\thepage}
}

\def\mysubject{}
\def\mykeywords{; ; }

\hypersetup{ % sets Metadata of the pdf file
pdftitle={\mytitle},
pdfsubject={\mysubject},
pdfauthor={\myauthor},
pdfkeywords={\mykeywords}
}

\begin{document}

\maketitle
\thispagestyle{firststyle}
\pagestyle{elsestyle}

%%%%%%%%%%%%%%%%%%%%%%%%%%%%%%%%%%%%%%%%%%%%%%%%%%%%%%%%%%%%%%%%%%%%%%%
%%%%%%%%%%%%%%%%%%%%%%%%%%%%%%%% ABSTRACT %%%%%%%%%%%%%%%%%%%%%%%%%%%%%
%%%%%%%%%%%%%%%%%%%%%%%%%%%%%%%%%%%%%%%%%%%%%%%%%%%%%%%%%%%%%%%%%%%%%%%
\begin{abstract}
%Automation in the life science research laboratory is a paradigm that has gained increasing relevance in recent years. Current robotic solutions have limited capacity for the transportation and manipulation of objects, thus preventing the realization of complex workflows. In this paper, we therefore present fingers for a parallel gripper for biolaboratory automation that can handle a wide range of liquid containers. We deduce a taxonomy of containers, based on which the need to develop a novel, flexible grasping solution follows. To achieve this flexibility, our fingers are developed as a monolithic dual-extrusion 3D print. The integration of a rigid and a soft material as well as the fingertip design are key features that enhance grasping capabilities. By adopting a passive compliant mechanism, a simple actuation system and a low weight are maintained. The ability to resist chemicals and high temperatures and the integration with a tool exchange system render the fingers usable for daily laboratory use and complex workflows. We present the task suitability of the fingers in experiments that show the wide range of vessels that can be handled as well as their tolerance against displacements and their grasp stability.
Automation in the life science research laboratory is a paradigm that has gained increasing relevance in recent years. Current robotic solutions often have a limited scope, which reduces their acceptance and prevents the realization of complex workflows. The transport and manipulation of laboratory supplies with a robot is a particular case where this limitation manifests. In this paper, we deduce a taxonomy of biolaboratory liquid containers that clarifies the need for a flexible grasping solution. Using the taxonomy as a guideline, we design fingers for a parallel robotic gripper which are developed with a monolithic dual-extrusion 3D print that integrates rigid and soft materials to optimize  gripping properties. We design fine-tuned fingertips that provide stable grasps of the containers in question. A simple actuation system and a low weight are maintained by adopting a passive compliant mechanism. The ability to resist chemicals and high temperatures and the integration with a tool exchange system render the fingers usable for daily laboratory use and complex workflows. We present the task suitability of the fingers in experiments that show the wide range of vessels that can be handled as well as their tolerance against displacements and their grasp stability. 
\end{abstract}

%%%%%%%%%%%%%%%%%%%%%%%%%%%%%%%%%%%%%%%%%%%%%%%%%%%%%%%%%%%%%%%%%%%%%%%
%%%%%%%%%%%%%%%%%%%%%%%%%%%% INTRODUCTION %%%%%%%%%%%%%%%%%%%%%%%%%%%%%
%%%%%%%%%%%%%%%%%%%%%%%%%%%%%%%%%%%%%%%%%%%%%%%%%%%%%%%%%%%%%%%%%%%%%%%
\section{INTRODUCTION}
\label{section:introduction}
Not least due to the COVID-19 pandemic, research in biological laboratories (biolabs) has recently gained scale and medial attention \cite{Nowakowska2020}. The use of robots that take over tasks in this endeavor has increased in recent years \cite{Courtney2021}, thus making it more reproducible, efficient, and safe \cite{Holland2020}. However, in particular in smaller labs, researchers often do not deploy robotic laboratory helpers due to their high costs and their low flexibility with respect to handling labware of different vendors and ever-changing laboratory protocols \cite{Holland2020}. If a team uses a robot despite these obstacles, the most common devices are liquid handling machines \cite{Kong2012}. However, the generic and efficient robotic execution of the preceding and subsequent steps, i.e., handling liquid containers and transporting them between different stations in the laboratory, is often neglected.

A typical process that involves handling many different containers is PCR preparation \cite{Elkins2013}: A sample is cultured in either a flask or a Petri dish, from which it is transferred to a microcentrifuge tube to extract its DNA. To this end, specific reagents are added that are stored in cryogenic tubes. The DNA is then pipetted into a 96-well plate and supplemented with ultrapure water, often kept in bottles of \SI{1}{\liter}. The variety of containers in this common process shows that using a separate gripping tool for each vessel is inefficient. Instead, relying on one gripper to handle all containers is advantageous, as is the natural case for a human.

Many systems for robotic biolaboratory automation employ 1-degree-of-freedom grippers. The relevant containers for simple workflows can be handled with a pinch grasp while maintaining a plain control mechanism. Liquid handling workstations, e.g., Beckman Coulter Biomek Cell Workstation \cite{Lehman2016}, often come with parallel grippers that can reliably move rectangular containers such as 96-well plates inside the device. This design limits the range of equipment that can be handled to the necessary minimum. The same applies to the Mobile Robotic Chemist \cite{Burger2020}, for which the authors use a parallel gripper with fine-tuned fingers to handle the labware in their experiment. Fleischer et al. describe a dual-arm robot performing different laboratory transport and manipulation tasks \cite{Fleischer2016}. They equip a parallel gripper with different fingers that depend on the main tasks of the arms. Although this approach allows for more flexibility, the range of tools is limited to two finger pairs simultaneously, and it requiring the availability of two robotic arms. 

\begin{figure}[!b]
    \centering
    \includegraphics[width=0.48\textwidth]{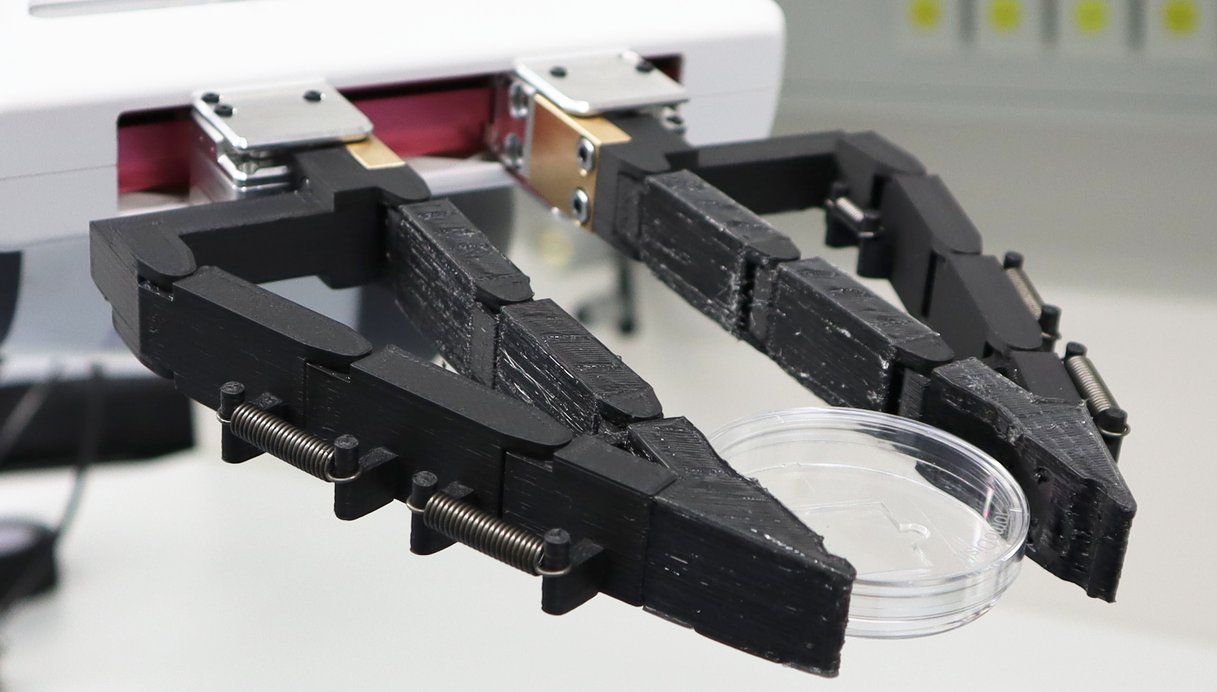}
    \caption{Fingers developed holding a Petri Dish.}
    \label{fig:gripper}
    \vspace{-10px}
\end{figure}

\tikzset{
  basic/.style  = {draw, drop shadow, rectangle, scale=1},
  level1/.style = {basic, thin,align=center, fill=white},
  level1inactive/.style={basic, thin,align=center,  fill=gray!80},
}
\begin{figure*}[!bh]
\centering
\scalebox{0.8}{
\begin{tikzpicture}[
    root/.style={sibling distance = 10cm, level distance = 1.1cm, basic, rounded corners=8pt, fill=white},
    level 1/.style={sibling distance = 10cm, level distance = 1.1cm, basic},
    level 1 inactive/.style={sibling distance = 10cm, level distance = 1.1cm, basic, fill=gray!80},
    level 2/.style={sibling distance = 5cm},
    level 3/.style={sibling distance = 3cm},
    {edge from parent fork down},
  ]
%root of the the initial tree, level 1
\node[root]{\textbf{Liquid Containers}} 
% The first level, as children of the initial tree
  child {node[level1] (c0) [xshift = -1.1cm] {\textbf{Lightweight  ($<$ \SI{300}{\gram}) - Pinch Grasp}}
    child {node[level1] (c00)[xshift = -0.4cm] {\textbf{Thin (Width $\leq$ \SI{1}{\centi\meter})}}
      child {node[level1inactive] (c000) [xshift = 0.1cm] {\textbf{Snap Cap}}
        child {node (c0001) {\textbf{1$\vert$Microcentrifuge Tubes} }}       }
      child {node[level1inactive] (c001) [xshift = -0cm] {\textbf{Screw Cap}}
        child {node (c0010) {\textbf{2$\vert$Cryotubes}}}
      }
    }
    child {node[level1] (c01) {\textbf{Medium (Width \SIrange{1}{8}{\centi\meter})}}
      child {node[level1inactive] (c011) [xshift = 0.1cm] {\textbf{Screw Cap}}
        child {node (c0101) {\textbf{3$\vert$Centrifuge Tubes} }
        }
      }
      child {node[level1] (c010) [xshift = -0.2cm] {\textbf{Loose Lid}}
        child {node (c0100) [align=center] {\textbf{4$\vert$Petri Dishes}}
        }
      }
    }
    child {node[level1] (c02) [xshift = 0.75cm] {\textbf{Wide (Width $\geq$ \SI{8}{\centi\meter})}}
      child {node[level1inactive] (c020) {\textbf{Screw Cap}}
        child {node (c0200) [align=center] {\textbf{5$\vert$Cell Culture Flasks} }
        }
      }child {node[level1inactive] (c021) [xshift = 0cm] {\textbf{Sealing Foil}}
        child {node (c0210) {\textbf{6$\vert$Well Plates} }
        }
      }
    }
  }
  child {node[level1] (c1) [xshift = -1.5cm] {\textbf{Heavyweight (\SIrange{300}{1500}{\gram}) - Power Grasp}}
    % child {node[level1] (c10) {\textbf{Enclosing Grasp}}
    child {
      child {node[level1inactive] (c100) [xshift = 1.6cm] {\textbf{Snap Cap}}
        child {node (c1000) {\textbf{7$\vert$Disinfectant} }} 
      }
      child {node[level1inactive] (c101) [xshift = 1.1cm] {\textbf{Screw Cap}}
        child {node (c1010) {\textbf{8$\vert$Glass Bottles} }} 
      }
    }
  };
\end{tikzpicture}
}%\vspace{-6pt}
    \caption{Requirement-based taxonomy of biolab liquid containers (see Fig.~\ref{fig:containers} for each corresponding container type). Opening and closing lids with grey background can be done using other devices.}
    % \vspace{-20pt}
    \label{fig:liquid_containers_taxonomy}
\end{figure*}

To increase the versatility of a gripper while maintaining a simple control mechanism, researchers have developed various methods \cite{Piazza2019}. Underactuation is a principle that enables a robotic system to have more degrees of freedom than actuators \cite{He2019}. Birglen has presented underactuated fingers called ``PaCoMe'' (\emph{Passive Cord Mechanism}) for a 1-degree-of-freedom parallel gripper that can grasp a variety of objects using pinch and power grasps \cite{Birglen2015}. Designed for industrial applications, these fingers are unsuitable for handling specific biolab labware. %Notable examples are small tubes that they cannot grip with the required robustness, or Petri dishes that they cannot lift at all because they would only grasp the overhanging lid. 
Anthropomorphic hands are a versatile class of end-effectors often based on underactuation (e.g., \cite{ParkH2020}); however, they still come with a higher number of actuators. The electromechanical control methods they require introduce another layer of complexity, thus rendering them impractical for biolab applications at their current developmental state.

Soft robotics fundamentally relies on underactuation: viscoelastic materials allow the continuum deformation of mechanical structures and generate virtually infinite degrees of freedom, enabling soft grippers to grasp various items \cite{Shintake2018}. 
One example is the utilization of the Fin Ray Effect, which is grounded on the physiology of fish fins, to maximize the contact area and thus contact forces \cite{Crooks2016}. Tendon-driven approaches are another variant that combine both rigid and soft materials and can therefore achieve pinch and power grasps \cite{Hussain2020}. However, a drawback of these  solutions is their inability to sense force feedback from the environment \cite{Culha2017}, which is a requirement in force-sensitive and highly precise manipulation tasks in a biolab.

Commercial grippers, such as the BarrettHand\footnote{\url{https://advanced.barrett.com/barretthand}, last accessed 13 February 2023} (Barrett Technology, USA) or the Robotiq 2F\footnote{\label{fn:robotiq}\url{https://robotiq.com/products/2f85-140-adaptive-robot-gripper}, last accessed 13 February 2023} (Robotiq, Canada), can manipulate a wide range of objects. However, their elaborate electromechanical control methods would introduce another layer of complexity into the laboratory. Consequently, researchers who want to employ these tools for complex workflows such as the PCR example described above, spend more time maintaining or extending the gripper.

% To address the shortcomings of the solutions presented regarding robotic biolab automation, we have developed a novel finger system for a parallel gripper. By designing it as a seamless 3D print for a dual extrusion printer, its properties could be tested and adjusted as part of a rapid prototyping process. The fine-tuned tip can handle a wide range of laboratory containers while providing good force transmission for sensitive and intelligent manipulation tasks. Employing a soft filament using the second extruder is an essential innovation that improves grasping stability, thus combining the advantages of hard and flexible materials. Underactuation as a passive mechanism allows heavy objects to be transported while maintaining a simple actuation system and a lightweight design. Eventually, the integration with a tool exchange system and resistance to high temperatures and chemicals allow for the application of the system in complex biolaboratory workflows.

To address the shortcomings of the solutions presented, we developed fingers that allow flexibility for grasping while keeping a simple integrability with a robotic system. We initially deduced a taxonomy of containers used in a biolaboratory to define "flexibility`` in this context. Containers are separated by weight, size, and opening, thus leading to principal requirements that a gripper must satisfy. Our contributions are in detail:
\begin{itemize}
    \item specification of a taxonomy of liquid containers used in a life science laboratory to deduce gripper requirements,
    \item identification and adjustment of suitable finger design to make it biolab-compatible,
    \item development of suitable fingertips based on the objects to be grasped,
    % item Monolithic dual-extrusion 3D print of the finger including a soft material to enhance grasping capabilities
    \item presentation of finger applicability in biolab tasks, including integration with a finger exchange system. 
\end{itemize}

%old: The remainder of the paper is organized as follows: We first establish a taxonomy of biolab liquid containers (Section~\ref{subsection:taxonomy}) and describe the robotic system that is used in our laboratory (Section~\ref{sec:robot}). Based on this setup, we derive requirements that a gripper must satisfy (Section~\ref{section:requirements}) and show in how far these are met by the state of the art presented above (Section~\ref{sec:requirement_fulfillment}). In Section~\ref{section:design}, we deduce the design of the fingers based on these requirements: We start with basic design considerations in Section~\ref{section:basic_considerations} and explain the passive compliant mechanism that we eventually adopt in Section~\ref{sec:passive_mechanism}. Section~\ref{sec:soft_material} deals with the choice of the soft material that we use in our 3D print. Both the design of the fingertips (Section~\ref{sec:tip_design}) and the joints (Section~\ref{sec:joint_design}) is then discussed in more detail. Section~\ref{sec:final_design} gives an overview of the final finger design, before the manufacturing with a 3D printer is described in Section~\ref{sec:print}. The usability of our fingers is validated in Section~\ref{section:experimental_results} by presenting the results of several qualitative and quantitative experiments and by benchmarking it against another finger pair. Finally, we provide an outlook in Section~\ref{section:outlook}.
The remainder of the paper is organized as follows: In Section~\ref{section:req_analysis}, we analyze the requirements that a suitable gripper must satisfy. Section~\ref{section:design} deals with the design based on these requirements and the fabrication of the gripper fingers using a 3D printer. Their usability is validated in Section~\ref{section:experimental_results} by presenting the results of several qualitative and quantitative experiments and benchmarking them against another finger pair. Finally, we summarize our findings and provide an outlook in Section~\ref{section:outlook}.

%%%%%%%%%%%%%%%%%%%%%%%%%%%%%%%%%%%%%%%%%%%%%%%%%%%%%%%%%%%%%%%%%%%%%%%
%%%%%%%%%%%%%%% PROBLEM STATEMENT AND CONTRIBUTION %%%%%%%%%%%%%%%%%%%%
%%%%%%%%%%%%%%%%%%%%%%%%%%%%%%%%%%%%%%%%%%%%%%%%%%%%%%%%%%%%%%%%%%%%%%%

\section{REQUIREMENT ANALYSIS}
\label{section:req_analysis}
\subsection{Taxonomy of Biolaboratory Containers}
\label{subsection:taxonomy}
\begin{figure}[!t]
    \centering
    \vspace{3px}\includegraphics[width=0.48\textwidth]{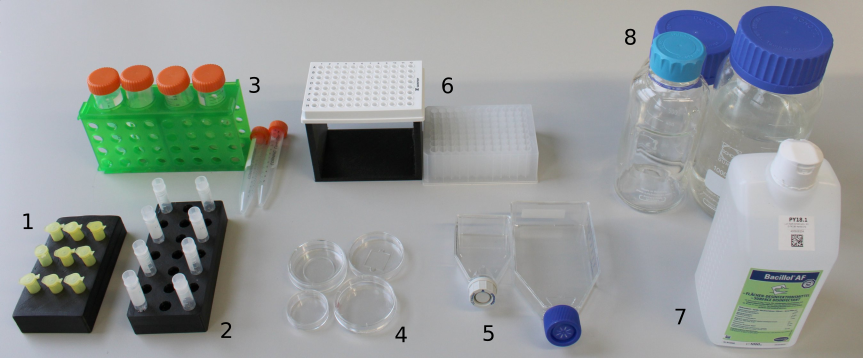}
    \caption{Variety of biolab liquid containers with numbers according to the taxonomy in Fig.~\ref{fig:liquid_containers_taxonomy}.}\vspace{-5px}
    \label{fig:containers}
\end{figure}

To define a gripper's requirements, we initially compiled a list of liquid containers regularly used in a biolab. We categorized them in a hierarchy based on their grasping-related properties (cf. Fig.~\ref{fig:liquid_containers_taxonomy}): We first separated them based on their weight, thus differentiating between lightweight items that can be precisely placed using a pinch grasp (up to \SI{300}{\gram} with liquid), and heavy equipment transported using a power grasp (up to \SI{1500}{\gram} with liquid). The latter category consists of bottles that are heavy due to the weight of their material and the amount of liquid contained. For pinch grasps, we differentiated further by the width and diameter, respectively, of the items in question: Different types of tubes are the thinnest containers that are handled. Petri dishes, in particular, belong to medium-sized vessels. Cell culture flasks and well plates with widths greater than \SI{8}{\centi\meter} are the widest items that must be handled. For further manipulation, we eventually differentiated by the closing mechanism of a vessel: Cryo- and centrifuge tubes, cell culture flasks, and many heavy bottles are closed with a screw cap. Microcentrifuge or PCR tubes and other bottles come with a snap cap. Petri dishes have a loose lid on top, whereas well plates are either open or eventually sealed with foil. The last row shows exemplary containers, as depicted in Fig.~\ref{fig:containers}.

We note here that this taxonomy might not be comprehensive because individual biolab tasks could require different containers or variants of those described here. This rule applies especially to the wide variety of commercially bottled liquids, where we consider the disinfectant as one example. However, particularly for the pinch grasp scenario, our taxonomy shows the vessels that are used for the most relevant biolab tasks (see e.g., \cite{Lavrentieva2018} for cell culture, or overviews provided by plasticware suppliers\footnote{e.g.,~\url{https://thermofisher.com/de/de/home/life-science/lab-plasticware-supplies.html}, last accessed 13 February 2023}) or variants of the latter, e.g., PCR tubes that are miniature versions of microcentrifuge tubes with smaller volumes (\SI{0.1}{\milli\liter} instead of \SI{1.5}{\milli\liter}). In addition, some containers come in standardized dimensions (see e.g. \cite{ANSI2011_1, ANSI2011_2} for 96-well plates).

Different grasp types have to be applied depending on the equipment used with a container, which is similar to the manufacturing grasp taxonomy presented by Cutkosky \cite{cutkosky1989}. We show a mapping of containers and grasps in Table~\ref{tab:grasps} and exemplary situations in Fig.~\ref{fig:grasping_scenarios}. 

\begin{table}
    \caption[Grasp table]{Grasp types (see Fig.~\ref{fig:grasping_scenarios}) per container (see Fig.~\ref{fig:liquid_containers_taxonomy})}\vspace{-10px}
    \label{tab:grasps}
    \begin{center}
    \begin{tabular}{l|cccc}
    \toprule
   Container $\downarrow$ / Grasp $\rightarrow$ &Top&Side&Insertion&Power\\
    \midrule
    1 $\vert$ Microcentrifuge tubes&\checkmark&\checkmark&&\\
    2 $\vert$ Cryotubes&\checkmark&\checkmark&&\\
    3 $\vert$ Centrifuge tubes&\checkmark&\checkmark&&\\
    4 $\vert$ Petri dishes&&\checkmark&&\\
    5 $\vert$ Cell culture flasks&\checkmark&\checkmark&&\\
    6 $\vert$ Well plates&\checkmark&&\checkmark&\\
    7 $\vert$ Disinfectant&&&&\checkmark\\
    8 $\vert$ Glass bottles&&&&\checkmark\\
    \bottomrule
    \end{tabular}\setlength\tabcolsep{6 pt}
    \end{center}
    \vspace{-20pt}
\end{table}

\subsection{Robotic Arm and Tool Exchange System}
\label{sec:robot}
We use a Franka Emika Research (Franka Emika GmbH, Germany) robotic arm in our experiments \cite{Haddadin2022}, equipped with the Franka Emika Hand\footnote{\label{fn:franka_hand}\url{https://franka.de/ecosystem/}, last accessed 28 February 2023}, a two-finger parallel gripper with one actuator. As the travel width of the gripper jaws is \SI{83}{\milli\meter}, an additional  set of requirements results from the robotic system. We employ a tool exchange system described in \cite{Ringwald2023} to enable the robot to exchange fingers autonomously.

\subsection{Requirement Derivation}
\label{section:requirements}
We can deduce the minimum requirements on a biolab gripper based on the container taxonomy in Section~\ref{subsection:taxonomy}: We require that a gripper can grasp a standard 96-well plate with a top width of \SI{83}{\milli\meter} \cite{ANSI2011_1, ANSI2011_2} (requirement R1a). At the other end of the scale, it must perform stable grasps of thin containers, such as microcentrifuge and cryotubes that measure down to \SI{6}{\milli\meter} in diameter (requirement R1b). Of the four closing mechanisms listed, the gripper is only required to handle the loose, overhanging lid. Devices to open screw caps are readily available on the market because of their omnipresence \cite{Jaeger2021}. The same applies to sealing 96-well plates, which has very different requirements that special sealing devices can meet. Snap caps can be found in different varieties (see Fig.~\ref{fig:containers}), and fixtures in the laboratory can be used to open them. However, handling containers with loose and overhanging lids requires our gripper to be designed in such a way that it can grasp them below the overhang. We call this requirement "tip suitability``, as it is used for side grasps with the tip, in particular for Petri dishes (cf. Table~\ref{tab:grasps}) (requirement R2). The third requirement derived from the taxonomy is that the gripper can carry containers up to \SI{1500}{\gram} (requirement R3).

The robotic system with which the gripper is used imposes further requirements on its design: As described in Section~\ref{section:introduction}, it is standard in laboratory automation today to use robots with a 1-degree-of-freedom parallel gripper, which we adopt here. For a robotic system to act as a productive and versatile helper in the biolab, it must be able to use different tools in one workflow, e.g., by handling a pipette as presented in \cite{Knobbe2022} after provisioning a container. If a gripper cannot handle these tools itself, it is, therefore, another requirement that its fingers can be changed autonomously using a tool exchange system. Likewise, complex robotic workflows often necessitate integrating force feedback from the environment into task execution. Hence, we require good force transmission and a lightweight gripper design. We summarize these requirements as "Robotic Suitability`` (requirement R4).

As a last requirement, we demand that the gripper withstand heat and treatment with chemicals (requirement R5) because biolab equipment is often in direct contact with biologically active materials, e.g., cells, and can thus be contaminated. For sterilization, labware is inserted into an autoclave that creates pressurized steam at \SI{121}{\celsius}. Therefore, to be reusable, a gripper must be able to withstand this treatment for \SIrange{30}{60}{\min} \cite{Lavrentieva2018}, as well as disinfection using ethanol or isopropyl alcohol.

\subsection{Requirement Fulfillment of the State of the Art}
\label{sec:requirement_fulfillment}
We conclude the requirements analysis by matching the state of the art described in Section~\ref{section:introduction} with them. The results are summarized in Table~\ref{tab:requirements} and show that no single solution satisfies all requirements R1-R5.

All of the grippers considered satisfy requirement R1, i.e. their strokes exceed our required range. This includes Burger's fine-tuned yet simple fingers \cite{Burger2020} if they are mounted to the Franka Hand gripper that we consider. We do not assume, however, that they can carry the maximum payload that we demand (R3) due to their low contact area. Birglen's underactuated fingers \cite{Birglen2015} and the Robotiq 2F\footref{fn:robotiq} have planar fingertip interfaces; therefore, they cannot handle containers with overhanging lids (R2). If they grasp a Petri dish, they would only lift the lid, but they cannot transport it with the vessel containing the liquid. The Robotiq gripper also lacks the robotic suitability (R4) that we demand, meaning that it cannot be used with a tool exchange system and it cannot handle pipettes or other tools. Finally, Birglen's fingers are 3D-printed with an undisclosed material "with properties very similar to ABS`` \cite{Birglen2015}. The latter starts deforming at a temperature of about \SI{100}{\celsius}\footnote{\url{https://3dinsider.com/melting-point-abs/}, last accessed 1 March 2023} and hence does not meet requirement R5. While the grip tape applied is not described in more detail, one can suppose that it is also not stable under these conditions.

\begin{table}%[!bt]
\caption[Requirement table]{Requirement fulfillment of the state of the art: Birglen \cite{Birglen2015}, Burger \cite{Burger2020}, Robotiq 2F\footref{fn:robotiq}}
\label{tab:requirements}\setlength\tabcolsep{4.5 pt}
\vspace{-10pt}
\begin{center}
\begin{tabular}{l|ccc}
\toprule
\hspace{-5px}Requirement&\hspace{-2px}Birglen &\hspace{-3px}Burger&\hspace{-3px}Robotiq\\
\midrule
\hspace{-5px}R1: Stroke exceeds $[\SI{5}{\milli\meter}; \SI{83}{\milli\meter}]$ & \checkmarksmall & \checkmarksmall & \checkmarksmall\\
\hspace{-5px}R2: Tip suitability & & \checkmarksmall & \\
\hspace{-5px}R3: Maximum payload $\geq\SI{1.5}{\kilogram}$ & \checkmarksmall & & \checkmarksmall\\
\hspace{-5px}R4: Robotic suitability & \checkmarksmall & \checkmarksmall & \\
\hspace{-5px}R5: Temperature and chemical resistant & & \checkmarksmall & \checkmarksmall \\

\bottomrule
\end{tabular}\setlength\tabcolsep{6 pt}
\end{center}
\vspace{-20pt}
\end{table}

%%%%%%%%%%%%%%%%%%%%%%%%%%%%%%%%%%%%%%%%%%%%%%%%%%%%%%%%%%%%%%%%%%%%%%%
%%%%%%%%%%%%%%%%%% Gripper Design and Production Steps %%%%%%%%%%%%%%%%%%%%%
%%%%%%%%%%%%%%%%%%%%%%%%%%%%%%%%%%%%%%%%%%%%%%%%%%%%%%%%%%%%%%%%%%%%%%%
\section{FINGER DESIGN AND FABRICATION}
\label{section:design}

\subsection{Basic Design Considerations}
\label{section:basic_considerations}
\begin{figure}[tb]
\centering
    \vspace{5px}\includegraphics[width=0.47\textwidth]{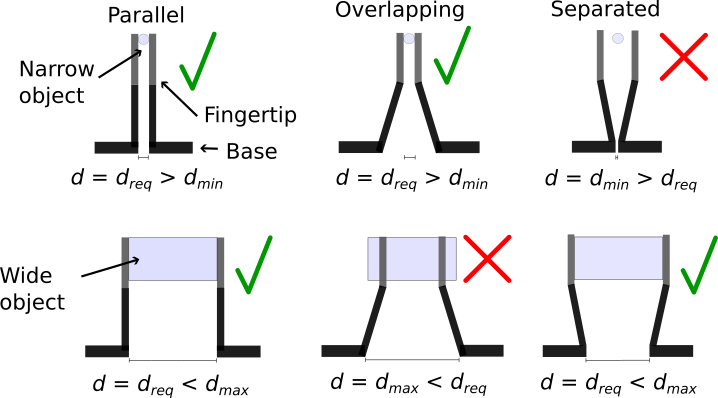}
   \caption{Pinch grasp: $d$, $d_{min}$, $d_{max}$, and $d_{req}$ are the displayed, minimum, maximum, and required opening widths of the gripper at the tip. Only fingertips parallel to the edge of the base allow grasping all objects (green checkmarks), while the overlapping and separated configurations do not (red crosses). }\vspace{-5px}
   \label{fig:narrow_wide_design}
\end{figure}

Comparing the minimum and maximum required opening widths ($\SI{5}{\milli\meter}$ and $\SI{83}{\milli\meter}$, respectively; requirement R1 according to Table~\ref{tab:requirements}) and the travel width of the gripper of $\SI{83}{\milli\meter}$ leads to restrictions on the position of the fingertips: They must be parallel and exactly touch each other when the gripper is closed. A narrow container could not be grasped with sufficient force if they were separated. If, conversely, the fingertips were overlapping too much, they would not be able to grasp a wide container. This circumstance is illustrated in Fig.~\ref{fig:narrow_wide_design} and prevents the use of e.g., the standard fingers that come with the Franka Hand\footref{fn:franka_hand}. We note here that many items are stored in racks; hence, they must be gripped with the tip of the fingertip when grasped from the top, as shown in Fig.~\ref{fig:grasping_scenario_eppi}. This prevents a design where the fingertip has a wide opening at the tip, followed by a step after which it gets narrow. The need to keep the gripper's design, fabrication, and control simple also prohibits using a gearbox that would allow for a larger transmission between the travel width of the jaws and the opening width at the tip.

\begin{figure}
\centering
    \includegraphics[width=0.47\textwidth]{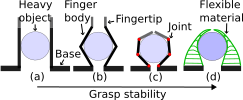}\vspace{-1px}
   \caption{Power grasp: The more to the right, the higher the grasp stability due to the bigger contact area of the finger. Shown designs: (a) Rigid, (b) rigid and bent, (c) rigid links connected with joints, (d) flexible material.}\vspace{-15px}
   \label{fig:body_design}
\end{figure}

These considerations only define the position of the fingertips with respect to the base but not the design of the fingers. The necessity to engineer this part, too, arises from the maximum transport payload (requirement~R3). A simple planar face as suggested by Fig.~\ref{fig:body_design}a results in a small contact area and low grasp stability, thus contradicting R3. A bent design improves the grasping properties (see Fig.~\ref{fig:body_design}b). However, this rigid design is not ideal for adapting to different shapes of the bottles to be expected. A flexible material, as presented in Section~\ref{section:introduction} and shown in Fig.~\ref{fig:body_design}d, excels in these tasks. Still, it has limitations in sensing feedback from the environment, as discussed in Section~\ref{section:introduction}. A suitable compromise is a semi-flexible design consisting of several rigid links connected with joints, thus allowing a finger to adjust better to the object's shape (see Fig.~\ref{fig:body_design}c).

%\vspace{-5px}

\subsection{Passive Compliant Mechanism}
\label{sec:passive_mechanism}
We based our design on a passive compliant mechanism to achieve flexibility for pinch, and power grasps using a parallel gripper with one actuator at the base. As described in Section~\ref{section:basic_considerations}, this choice allows high stability for power grasps while maintaining the ability to perform precision grasps with the tip. Birglen's underactuated ''PaCoMe``-fingers \cite{Birglen2015} meet these requirements, using three springs as passive elements to increase the degrees of freedom. We decided to adopt this design. However, as it fails to meet two of the five requirements that we posit, we need to introduce key modifications to make it useable in a biolab context.

\subsection{Choice of Soft Material}
\label{sec:soft_material}
We decided to fabricate the fingers as a monolithic dual-extrusion 3D print to pursue a rapid-prototyping approach and generate reproducible results. This allowed us to use a soft printing material for the tip and the two inside links, as shown in Fig.~\ref{fig:drawing}, to enhance the grasping capabilities of the fingers. This feature mimics the properties of soft human skin. By this means, the fingers adapt better to the shape of an object, dampen shocks at the moment of grasping, and the friction between the fingers and containers is increased. Form closure also guarantees the stability of the dual-material print. We discarded depending on grip tape, an often used simple and cheap option, such as seemingly for Birglen's fingers \cite{Birglen2015}. On the one hand, it does not adapt to the fine-grained design of our fingertip (cf. Section.~\ref{sec:tip_design}). On the other hand, it is typically not resistant to constant exposure to high temperatures and alcohol, i.e., it does not conform to requirement~R5.

Instead, we used FilaFlex Ultrasoft (Recreus Industries, S.L., Spain) for the second extruder to coat the finger elements with a soft material. This thermoplastic polyurethane (TPU) has Shore hardness A~70.\footnote{\url{https://smooth-on.com/pw/site/assets/files/30090/durometerchart.png} for a hardness scale, last accessed 1 March 2023} It is resistant to heat and chemicals in comparison to other common 3D-printable materials, e.g. NinjaTek Cheetah (Fenner Drives, Inc., USA). We also tried FilaFlex Pro (Shore hardness A~60; Recreus Industries, S.L., Spain), which is the softest filament available for a fused deposition modeling (FDM) printer like the one used by us (see Section~\ref{sec:print} for the printing process). We eventually decided against it because of its lower resistance to wear and tear experienced in our experiments. We also tested the slightly less flexible Arnitel ID 2045 (Shore hardness D~34; DSM, The Netherlands). While this material is well printable, its friction coefficient and shape adaptability lag behind FilaFlex Ultrasoft.

As an alternative to dual-extrusion printing, we designed a mold into which a finger made of plain PLA is inserted, and that is then filled with silicone (Mold Star 15 SLOW and Ecoflex 00-30, Smooth-On, Inc., USA; Shore hardness A~15 and 00-30, respectively) to create a sleeve around the tip. We found this unfavorable due to too high friction and low durability of these materials. Because of this and the time-consuming fabrication, we eventually chose the dual-extrusion printing approach.

\subsection{Fingertip Design}
\label{sec:tip_design}
\begin{figure}[!t]
  \centering
  \includegraphics[width=.49\textwidth]{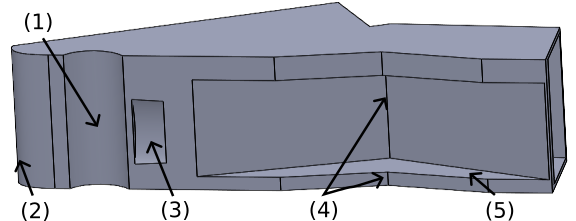}
  \caption{Fingertip features: (1): Cutout for tubes (side grasp), (2) rounded tip, (3) cutout for microcentrifuge tubes (top grasp), (4) cutouts for Petri dishes and centrifuge tubes and (5) overhang to grasp Petri dishes.}\vspace{-15px}
\label{fig:gripper_tip}
\end{figure}

A sketch of the fingertip surface printed with the soft material described in Section~\ref{sec:soft_material} is shown in Fig.~\ref{fig:gripper_tip}. To accommodate the multitude of shapes in the pinch scenario (cf. the taxonomy in Fig.~\ref{fig:liquid_containers_taxonomy}), we designed it with several cutouts. This feature enables the simultaneous grasp of a Petri dish and its lid, which is often wider than the container itself and lies loosely on top. It also provides stable grasps of other round containers, e.g., centrifuge tubes, which must be gripped firmly to decap them. Additionally, we slightly rounded the first millimeters of the tip so containers are ``guided'' into the opening. This feature is crucial because, according to the manufacturer, the maximum opening width of the gripper is \SI{83} {\milli\meter}, i.e., equal to the width of a 96-well plate. Including the fillet increased the tolerance against displacements while picking up wide objects.

\subsection{Joint Design and Spring Considerations}
\label{sec:joint_design}
\begin{figure}[!t]
  \centering
  \subfloat[][Joint with (1) rectangular stopper and (2) optimized gap between inner and outer part.]{\includegraphics[width=.23\textwidth]{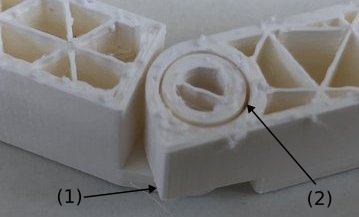}\label{fig:gripper_joint}}\quad
  \subfloat[][Finger holder with (1) latch bolt, (2) loosening hooks and (3) holder bars.]{\includegraphics[width=.23\textwidth]{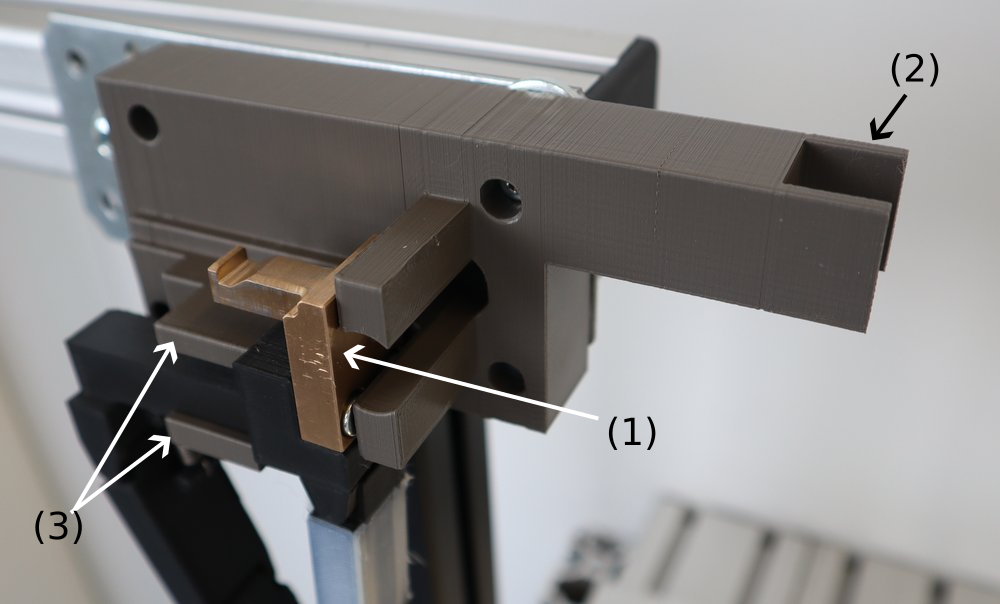}\label{fig:gripper_holder}}
  \caption{3D-printed features of the finger system}\vspace{-5px}
  \label{fig:finger_features}
\end{figure}

The joints connecting two links, as shown in Fig.~\ref{fig:gripper_joint}, are another feature of the monolithic print. To still allow them to rotate, the optimal gap distance between the inner and outer parts was determined. While a smaller gap leads to better finger stability and to less wobbly behavior, a too-small space makes both parts stick together and renders the print unusable. We found \SI{0.3}{\milli\meter} to be the distance that optimizes these constraints with our print settings (\SI{0.2}{\milli\meter} layer height, \SI{0.4}{\milli\meter} line width). To prevent overstretching in the wrong direction, we also added rectangular stoppers to the appropriate sides of the joints.

We chose the springs according to the theoretical derivation in \cite{Birglen2015}, i.e., a stiff spring close to the base (spring constant $D_\text{i}$, see Fig.~\ref{fig:drawing}) and two softer springs for the other two locations (spring constant $D_\text{o}$). Unlike the author, we placed the spring near the base inside to use tension springs here, while outside we would have needed to insert compression springs. As is visible in Fig.~\ref{fig:grasping_scenarios}, the placement inside does not limit the range of containers grasped. % Moreover, unlike Birglen's, our fingers remain stiff when pushed against the outside of the outer links, thus granting good force transmission.

\subsection{Final Design}
\label{sec:final_design}
\begin{figure}[!t]
  \centering
  \includegraphics[width=0.48\textwidth]{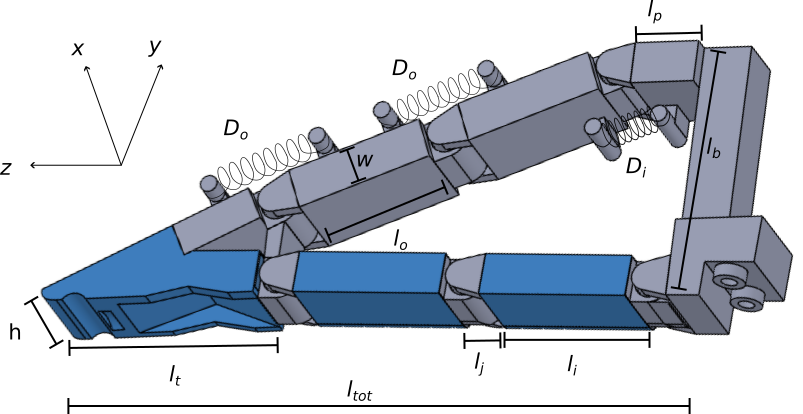}
  \caption{Drawing of a right finger. Grey: Printed with PLA. Blue: Printed with TPU.}
  \label{fig:drawing}
\end{figure}

We provide a drawing of a finger in Fig.~\ref{fig:drawing} and an overview of the measurements and other characteristics in Table~\ref{tab:gripper_stats}. We experimented with different lengths of the links and the base and concluded that dimensions similar to the ones presented in \cite{Birglen2015} are favorable. On the one hand, this allows grasping all containers presented in the taxonomy (Fig.~\ref{fig:liquid_containers_taxonomy}), particularly the wide bottles. On the other hand, even when the jaws are fully opened, the fingers only exceed the width of the Franka Hand gripper (visible in Fig.~\ref{fig:gripper} in the top left corner) by \SI{1}{\centi\meter}, i.e., we did not encounter limitations due to their size. Besides, adopting similar lengths as in \cite{Birglen2015} allows to transfer conclusions regarding the grasp stability analysis performed in that publication.

We believe that the fingers described in this study are scalable, e.g., their miniaturization can be a desirable feature for future developments, particularly in combination with a different gripper. However, a design study to optimize the sizes of all characteristics would exceed the scope of this publication.  

\begin{table}[!t]
\caption[Finger stats]{Finger characteristics}
\label{tab:gripper_stats}
\vspace{-8px}
\begin{center}
\begin{tabular}{c c}
\toprule
Feature & Size\\
\midrule
Length of inside links $l_\text{i}$&\SI{3.54}{\centi\meter}\\
Length of outside links $l_\text{o}$&\SI{3.20}{\centi\meter}\\
Joint length $l_\text{j}$&\SI{0.96}{\centi\meter}\\
Protrusion length $l_\text{p}$&\SI{1.52}{\centi\meter}\\
Tip length $l_\text{t}$&\SI{5.09}{\centi\meter}\\
Finger length from base to tip $l_\text{tot}$&\SI{15.05}{\centi\meter}\\
Finger height $h$&\SI{2.10}{\centi\meter}\\
Link width $w_\text{l}$ &\SI{1.00}{\centi\meter}\\
Base length $l_\text{b}$&\SI{6.45}{\centi\meter}\\
\midrule
Soft material layer width at tip inside&\SI{0.15}{\centi\meter}\\
Soft material layer width around links&\SI{0.02}{\centi\meter}\\
\midrule
Spring constant outside springs $D_\text{o}$&\SI{13.80}{\newton\per\centi\meter}\\
Spring constant of inside spring $D_\text{i}$&\SI{44.97}{\newton\per\centi\meter}\\
\midrule
Weight (after print)&\SI{48}{\gram}\\
Weight (with springs and latch bolt)&\SI{99}{\gram}\\
Price per print (without springs and latch bolt)&$\approx 3$~\euro\\
\bottomrule
\end{tabular}
\end{center}
\vspace{-20px}
\end{table}

\subsection{3D Print, Postprocessing and Tool Exchange}
\label{sec:print}
We developed our fingers in SolidWorks 2020 (Dassault Syst\`{e}mes, France), sliced in Cura~5.2 (Ultimaker, The Netherlands), and finally printed with an Ultimaker~3 Extended dual extrusion filament printer. We identified CHRONOS PLAtech (OLYMPfila, Germany) as the suitable primary material because it can resist temperatures of up to \SI{125}{\celsius} after post-tempering, as well as treatment with alcohol, thus meeting requirement R5.

The fabrication in one print limits the necessary post-processing to a minimum: The joints are loosened in case the parts stick slightly together, and lubricant is added. The grasping surfaces are also coated with three layers of rubber spray (mibenco GmbH, Germany) to increase friction further. According to the manufacturer, this spray is resistant to the required temperature and alcohol. Eventually, the springs and the latch bolt to mount the fingers to the gripper are attached.

Fig.~\ref{fig:gripper_holder} shows the finger holder designed for the finger in combination with the tool exchange system described in \cite{Ringwald2023}. It is mounted fixed and keeps a finger while not in use. The robot moves the jaw against the latch bolt (1) to pick up a finger. With sufficient force, the bolt slips into place, and a spring in the jaw secures it. To place the finger down again, the jaw is first moved against the two hooks on the right (2) that loosen the spring and release the finger. The latter is then placed again between the two bars (3).

%%%%%%%%%%%%%%%%%%%%%%%%%%%%%%%%%%%%%%%%%%%%%%%%%%%%%%%%%%%%%%%%%%%%%%%
%%%%%%%%%%%%%%%%%%%% EXPERIMENTAL RESULTS %%%%%%%%%%%%%%%%%%%%%%%%%%%%%
%%%%%%%%%%%%%%%%%%%%%%%%%%%%%%%%%%%%%%%%%%%%%%%%%%%%%%%%%%%%%%%%%%%%%%%
\section{EXPERIMENTAL RESULTS}
\label{section:experimental_results}

\subsection{Static Grasping Capabilities}
\label{sec:grasping_capabilities}
\begin{figure}[!tb]
  \centering
  \subfloat[][Microcentrifuge tube top grasp\label{fig:grasping_scenario_eppi}]{\includegraphics[width=.23\textwidth]{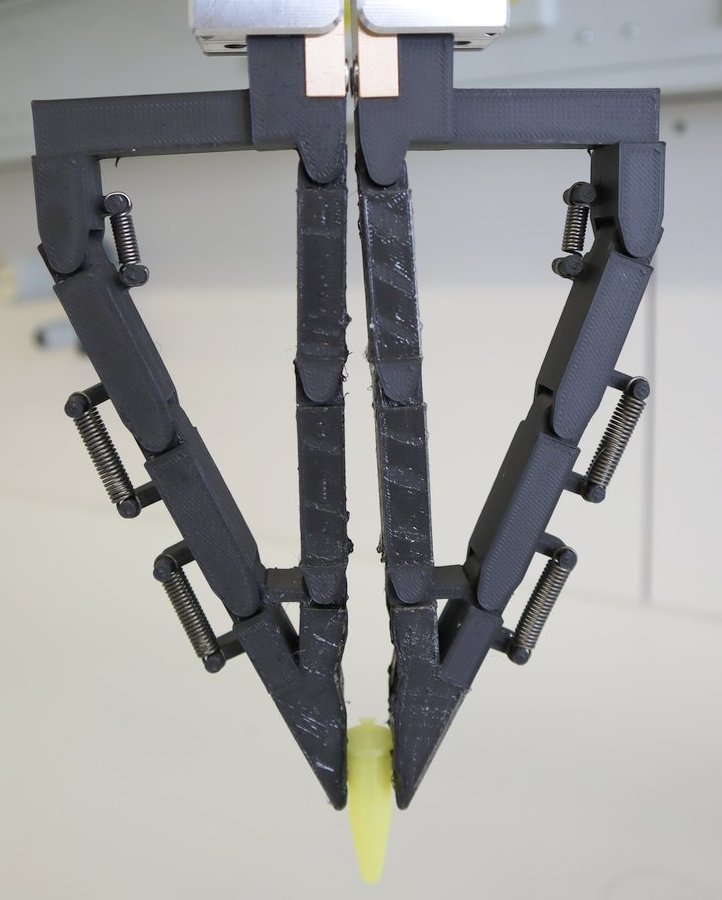}}  \quad
    \subfloat[][Duran bottle power grasp]{\includegraphics[width=.23\textwidth]{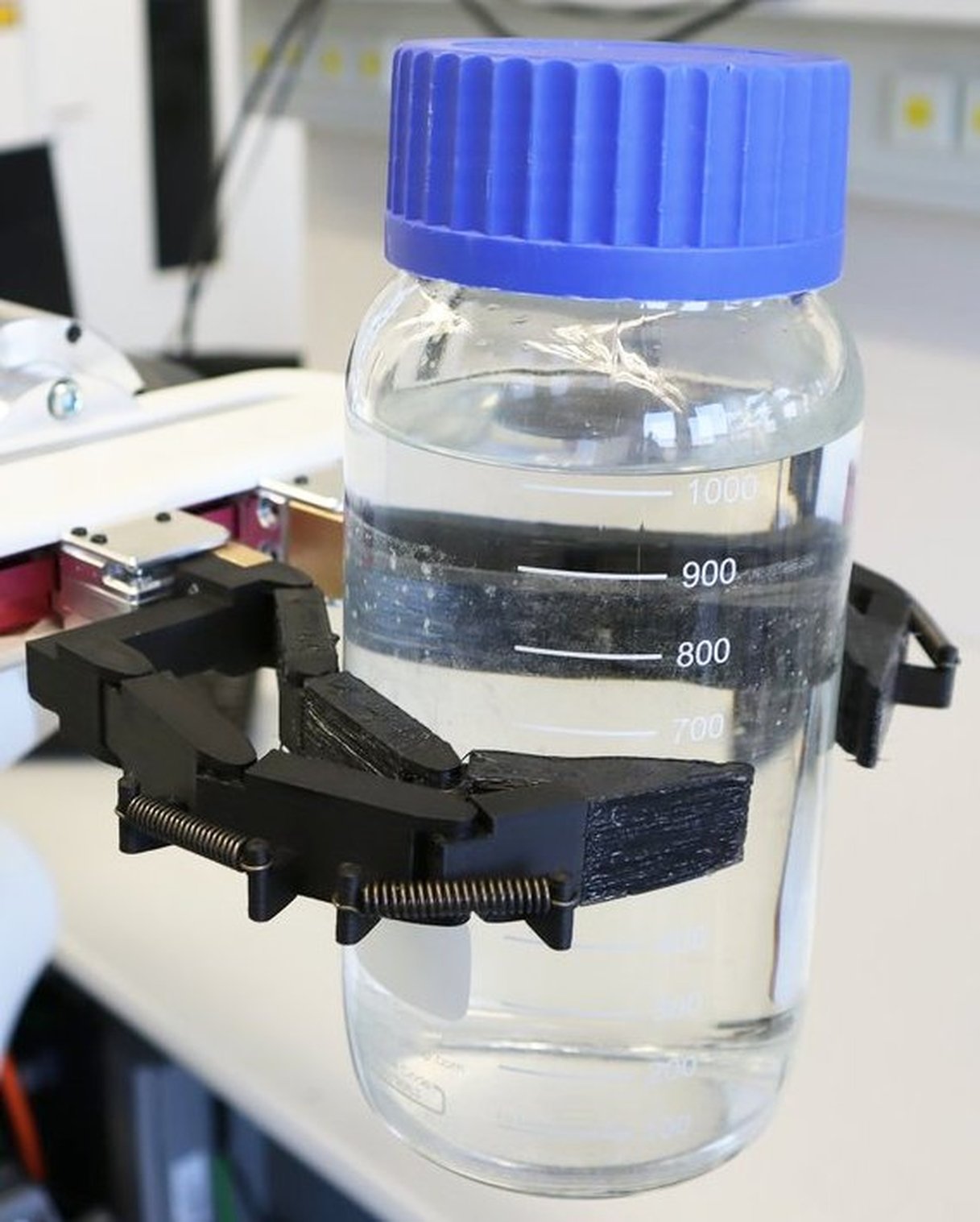}} \quad
    %\subfloat[][96-well plate top grasp]{\vspace{-10px}\includegraphics[width=.23\textwidth]{20220903_wellplate_top.JPG}}
  \\
    \subfloat[][Cell culture flask side grasp]{\includegraphics[width=.23\textwidth]{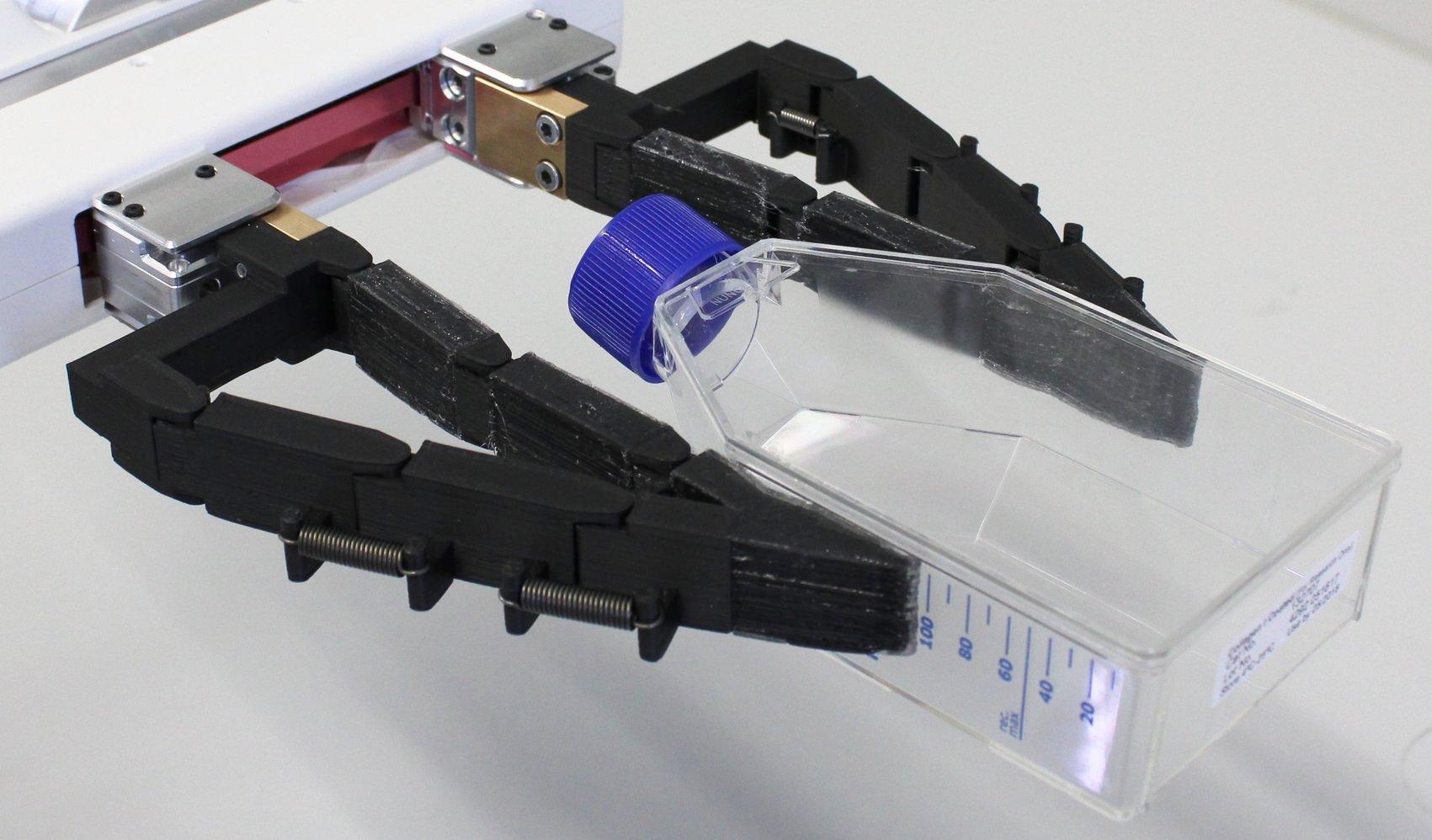}}\quad
    \subfloat[][96-well plate insertion grasp]{\includegraphics[width=.23\textwidth]{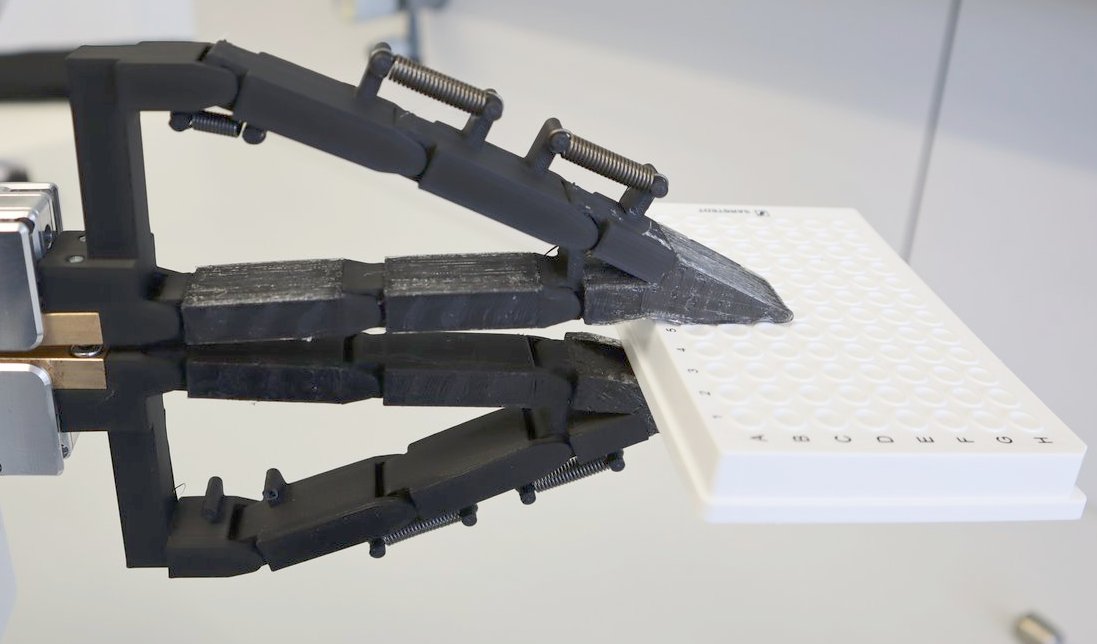}\label{fig:wellplate_side}}
  %\\
  %\subfloat[][Cell culture flask side grasp]{\includegraphics[width=.23\textwidth]{20220908_flask_side.JPG}}\quad
  %\subfloat[][Duran bottle power grasp  ]{\includegraphics[width=.23\textwidth]{20220903_bottle.JPG}}
  \caption{Grasping scenarios}\vspace{-15px}
  \label{fig:grasping_scenarios}
\end{figure}

Various static grasping scenarios are shown in Figs.~\ref{fig:gripper} and~\ref{fig:grasping_scenarios}. We note that all relevant containers from the taxonomy can be grasped and lifted with the grasp type combinations defined in Table~\ref{tab:grasps}. For some containers, particularly for tubes where some processes require the subsequent handling of several objects, we additionally designed 3D-printed storage racks with suitable distances between the individual containers that allow clean grasps. 

% \begin{figure*}
%     \begin{center}
%     \includegraphics[width=\textwidth]{20230623_grasp_experiments.png}
%     \vspace{-20px}
%     \caption{Finger performance measurements. See Fig.~\ref{fig:containers} for containers and Fig.~\ref{fig:grasping_scenarios} for grasp types. Directions conform to end-effector frame (see Fig.~\ref{fig:drawing}). (a) Maximum displacement tolerance for our fingers. Values for items with * depend on storage rack dimensions and spacings. (b) Force robustness for our fingers. (c) Force robustness for benchmark fingers.}
%     \label{fig:displacements}\vspace{-15px}
%     \end{center}
% \end{figure*}

\begin{figure*}
    \begin{center}
    \includegraphics[width=\textwidth]{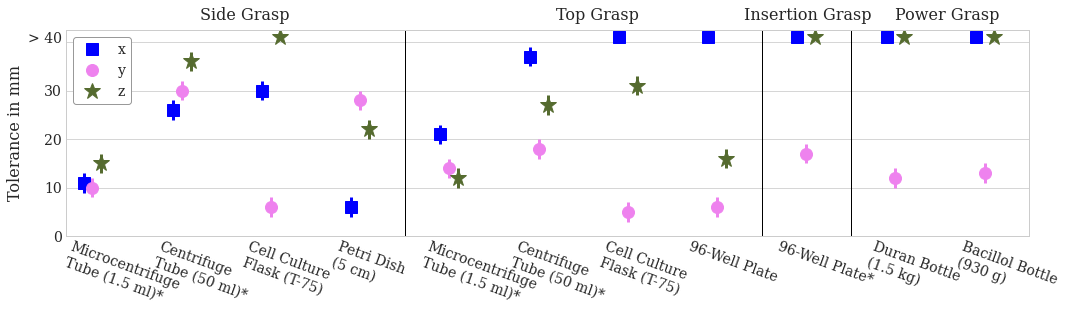}
    \vspace{-20px}
    \caption{Maximum displacement tolerance measurements. See Fig.~\ref{fig:containers} for containers and Fig.~\ref{fig:grasping_scenarios} for grasp types. Directions conform to end-effector frame (see Fig.~\ref{fig:drawing}).  Values for items with * depend on storage rack dimensions and spacings. Error bars denote the positioning accuracy of \SI{2}{\milli\meter} on grid paper.}
    \label{fig:displacements}\vspace{-15px}
    \end{center}
\end{figure*}
\begin{figure*}
    \begin{center}
    \includegraphics[width=\textwidth]{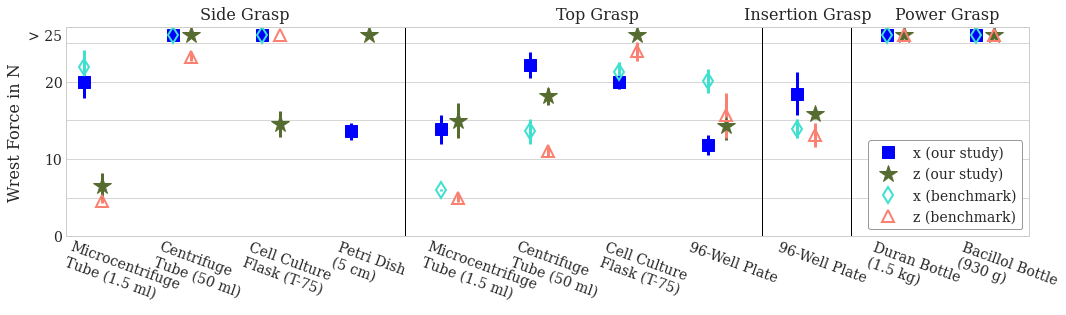}
    \vspace{-20px}
    \caption{Grasp robustness measurements for our fingers and benchmark. See Fig.~\ref{fig:containers} for containers and Fig.~\ref{fig:grasping_scenarios} for grasp types. Directions conform to end-effector frame (see Fig.~\ref{fig:drawing}). The benchmark cannot grasp the Petri dish, hence it is not shown. Error bars display the standard deviation from five measurements.}
    \label{fig:forces}\vspace{-22px}
    \end{center}
\end{figure*}

\subsection{Container Displacement Tolerance}
\label{sec:displacements}
To test the tolerance of our fingers with respect to small displacements of a container during grasping, we measured the maximum distance that the latter may have from an ideal position while still ensuring a stable grasp. This happened along all three dimensions for the various grasp types shown in Fig.~\ref{fig:grasping_scenarios}. We use the end-effector frame to describe the directions, i.e., $z$ points along the longest finger side, $y$ along the base and the sliding direction of the jaws, and $x$ along the finger height (cf. Fig~\ref{fig:drawing}).

% \begin{figure*}[!b]
%    \begin{center}
%    \includegraphics[width=\textwidth]{20230224_grasp_benchmark.png}
%    \vspace{-20px}
%    \caption{Force robustness plot for benchmark PaCoMe fingers. See Fig.~\ref{fig:containers} for containers and Fig.~\ref{fig:grasping_scenarios} for grasp types.}
%    \label{fig:benchmark}\vspace{-15px}
%    \end{center}
%\end{figure*}

We manually placed the container at a fixed position on grid paper with \SI{1}{\milli\meter} spacing to measure the tolerance. For each axis, we positioned the end-effector in an optimal position above the object and instructed it to be shifted a certain distance before grasping. This happened using the robot control software Desk (Franka Emika GmbH, Germany) that allows \SI{1}{\milli\meter} accuracy. The resulting distance is the sum of the maximum displacements in positive and negative direction that still facilitate a stable grasp. Reporting this way also accounts for asymmetric objects. The accuracy thus achieved and shown in the plot is, therefore, \SI{2}{\milli\meter}.

The results are shown in Fig.~\ref{fig:displacements}a. We chose containers from all groups as presented in Fig.~\ref{fig:liquid_containers_taxonomy} except the cryotubes, which have roughly the same dimensions as microcentrifuge tubes. Values bigger than \SI{40}{\milli\meter} are denoted as "$>40$``, implying that the configuration in question is very stable with respect to displacements. We see that the tolerance in $y$-direction is low for cases where the object size is close to the gripper's minimum or maximum opening width (e.g., 96-well plates). This is directly related to requirement R1, where we demand that the stroke of the gripper must span and exceed the whole range of the container diameters. The $x$-direction tolerance for the Petri dish is low due to its low height.

\subsection{Grasp Robustness}
\label{sec:force_robustness}
To test the robustness of a grasp, we measured at what applied force it becomes unstable. To this end, we used Franka Desk to grasp the same containers as in Section~\ref{sec:displacements} with grasp forces of \SI{20}{\newton} (pinch grasps) and \SI{100}{\newton} (power grasps), respectively. Then, a force gauge (FK 25, Sauter GmbH, Germany; \SI{25}{\newton} maximum force) was attached to each container with a wire. We pulled the gauge and noted down for the $x$- and $z$-direction in the end-effector frame at what force the container started moving or was wrested (i.e., the object cannot escape along the direction of the base ($y$-axis), so no measurements were taken along this direction). This process was repeated five times. The results are shown in Fig.~\ref{fig:forces}, where the error bars depict the standard deviation of the five measurements. Values exceeding the gauge range are very stable and shown as "$>25$``. One can see that the grasps with a large contact area, such as power grasps, or those relying on a form closure, e.g., side grasps for centrifuge tubes, are very stable. The only combination with a force below \SI{10}{\newton} that should be optimized is the $z$-direction for microcentrifuge tubes.
\addtolength{\textheight}{-3mm}

\subsection{Benchmark against PaCoMe-like Fingers}
To benchmark the fingers we developed based on our taxonomy, we compared their performance in the force robustness measurements against fingers resembling Birglen's PaCoMe gripper \cite{Birglen2015}. To this end, we 3D-printed a version of the finger that adopts the dimensions from this publication as closely as possible. We also glued grip tape (Comfort Overgrip, Wilson Sporting Goods Company, USA) to all surfaces used for grasping. The results are shown in Fig.~\ref{fig:forces} alongside our fingers for a simple comparison. We do not compare the displacement tolerance of both finger pairs because the improvements we introduce concern only the grasp stability but not the geometric properties that are the determining quantities behind the displacement measurements.

The Petri dish cannot be lifted because a gripper with a planar interface as the PaCoMe fingers only grasp the loose lid. We can also see that the benchmark performs similarly to our development. In some cases, it performs even better, e.g., the side grasp for cell culture flasks, because the contact area for our fingers is decreased due to the cutouts we use. However, one must note that this is only the case for values higher than \SI{10}{\newton} anyway. This tradeoff is acceptable for our case as we do not aim to maximize grasp forces but instead require a stable transport. Contrastingly, for the top grasp of microcentrifuge tubes, the wrest force is significantly increased from \SI{5.0(5)}{\newton} to \SI{14.9(22)}{\newton}, i.e., this combination is notably more stable for our fingers.

\section{CONCLUSION}
\label{section:outlook}
% We presented an enhanced finger system for a parallel gripper for robotic biolab automation and validated its grasping capabilities. In the future, we intend to integrate the fingers into typical laboratory workflows to validate their functioning and features in actual experimental settings. This can occur as part of the Intelligent Robotic Lab Assistants, as presented in \cite{Knobbe2022}, where two arms are mounted on a gantry system. For better integration, special focus should be placed on the kinematic structure of the fingers, i.e. the miniaturization of their dimensions, to make them compatible with more delicate grasping processes. The long-term durability of the print under continuous exposure to forces during lifting, combined with strains from the sterilization process, must be validated. Complementing the hardware solution proposed so far with software tools that report failed grasping attempts or reduce sloshing will yield valuable feedback that can be integrated into the control loop. 

In this paper, we presented 3D-printed fingers for a robotic parallel gripper that can grasp a wide range of vessels in a biological laboratory. We provide a taxonomy of these containers that, combined with the robotic setup used, serves as a guideline for their design. The rapid-prototyping approach applied based on a dual-extrusion 3D print allowed us to test and optimize different design parameters, such as the fingertip and the soft material. Eventually, we presented both qualitative and quantitative examples of their performance in handling relevant containers. This contrasts the original PaCoMe-design in \cite{Birglen2015}, where the author focused on the theoretical derivation of grasping properties and confined himself to qualitative application examples aimed at industrial purposes. The benchmarking experiments show how the modifications we implemented improve the grasping capabilities by enabling the grasp of certain containers at all or with higher robustness.  

In the future, we intend to integrate the fingers into typical laboratory workflows to validate their functioning and features in long-term experiments. In particular, combined with Intelligent Robotic Lab Assistants \cite{Knobbe2022}, where two arms are mounted on a gantry system, this allows for complex workflows, including pipetting and sample transport along a lab bench. Complementing the hardware solution proposed here with software tools that report failed grasping attempts or reduce sloshing will yield valuable feedback that can be integrated into the control loop.

% A conclusion section is not required. Although a conclusion may review the main points of the paper, do not replicate the abstract as the conclusion. A conclusion might elaborate on the importance of the work or suggest applications and extensions. 

%\addtolength{\textheight}{-12cm}   % This command serves to balance the column lengths
                                  % on the last page of the document manually. It shortens
                                  % the textheight of the last page by a suitable amount.
                                  % This command does not take effect until the next page
                                  % so it should come on the page before the last. Make
                                  % sure that you do not shorten the textheight too much.

%%%%%%%%%%%%%%%%%%%%%%%%%%%%%%%%%%%%%%%%%%%%%%%%%%%%%%%%%%%%%%%%%%%%%%%%%%%%%%%%

%%%%%%%%%%%%%%%%%%%%%%%%%%%%%%%%%%%%%%%%%%%%%%%%%%%%%%%%%%%%%%%%%%%%%%%%%%%%%%%%
% \section*{APPENDIX}

% Appendixes should appear before the acknowledgment.
\section*{ACKNOWLEDGMENT}

Please note that S. Haddadin has a potential conflict of interest as a shareholder of Franka Emika GmbH.

%%%%%%%%%%%%%%%%%%%%%%%%%%%%%%%%%%%%%%%%%%%%%%%%%%%%%%%%%%%%%%%%%%%%%%%%%%%%%%%%

\bibliographystyle{myIEEEtran} %unsrt}
\bibliography{main.bib}
\end{document}